# A Neural Network Model for Determining the Success or Failure of High-tech Projects Development: A Case of Pharmaceutical industry

Hossein Sabzian[1] [*], Ehsan Kamrani[2], Seyyed Mostafa Seyyed Hashemi[3]

[1]Department of Progress Engineering, Iran University of Science and Technology, Tehran, Iran

[2] Bio-Optics Lab, Wellman Center for photomedicine Harvard-MIT Health Sciences and Technology, Harvard Medical School, Harvard University, Cambridge, MA, USA

[3]Tadbir Economic Development Group, Tehran, Iran

*hossein_sabzian@pgre.iust.ac.ir

## Abstract

Financing high-tech projects always entails a great deal of risk. The lack of a systematic method to pinpoint the risk of such projects has been recognized as one of the most salient barriers for evaluating them. So, in order to develop a mechanism for evaluating high-tech projects, an Artificial Neural Network (ANN) has been developed through this study. The structure of this paper encompasses four parts. The first part deals with introducing paper's whole body. The second part gives a literature review. The collection process of risk related variables and the process of developing a Risk Assessment Index system (RAIS) through Principal Component Analysis (PCA) are those issues that are discussed in the third part. The fourth part particularly deals with pharmaceutical industry. Finally, the fifth part has focused on developing an ANN for pattern recognition of failure or success of high-tech projects. Analysis of model's results and a final conclusion are also presented in this part.

**Keywords**: High-tech Project Risk, Pharmaceutical industry, Risk Assessment Index System (RAIS), Principal Component Analysis (PCA), Artificial Neural Network (ANN), Pattern Recognition.

# 1. Introduction

Development project of high-tech products is always influenced by several risks neglecting each of which will dramatically undermine the success rate of such a project. Likewise, because of the fact that investment on development projects of high-tech products require the utilization of different resources (i.e. both physical assets & intellectual capitals) and will not always result in desired predictions, failure of such projects will doubtlessly inflict massive economic costs on organizations. Therefore, if project planners are enabled to measure and analyze the risk of such projects, they can forecast their success or failure more confidently.

The purpose of this study is to construct a model by which project managers can forecast the final consequence of investing on high-tech products. Thus, it contributes largely to stop investing those projects which are more likely to fail with regard to organization's current resources. This model is formulated through two interrelated phases. In the first phase, a number of risk-related variables (of high-tech projects) are gleaned. Then, the Principal Component Analysis (PCA) is used for analyzing them in order to construct a Risk Assessment Index System (RAIS) for high-tech products development projects and the second phase deals with developing an Artificial Neural Network (ANN) for recognizing success and failure pattern of high-tech projects in a pharmaceutical industry.

# 2. Literature Review

In the field of Artificial Intelligence (AI), an Artificial Neural Network (ANN) is known as a powerful computational data model that is able to extract and represent nonlinear input/output relationships among variables (Somers and Casal, 2009) As stated in Neurosolutions (2014) "The motivation for the development of neural network technology stemmed from the desire to develop an artificial system that could perform "intelligent" tasks similar to those performed by the human brain ". ANNs are basically presented as systems of interconnected

"neurons" that are able to compute values from inputs, and have the capability of machine learning as well as pattern recognition because of their adaptive nature.

In real world problems, ANNs have been applied in a wide range of fields ranging from aerospace engineering to banking industry. Hakimpour *et al.* (2011) have conducted a research on ANNs' applications in management in which they have classified its applications based on three main areas and their related problem types. Table 1 shows this classification.

Regarding Table 1 which is adapted from Hakimpour *et al.* (2011), it can be seen that ANN has been widely used in various types of business problems. In terms of risk assessment of high-tech products, some researches have done good works. Wang *et al.* (2000) proposed a radial basis function neural network and applied it to the risk evaluation of high-technology project investment. Song *et al.* (2005) developed a discrete Hopfield neural network for evaluating the investment risk of high-tech projects. Jiang *et al.* (2010) designed an ANN for assessing investment risks on high-tech projects.

Badiru and Sieger (1998) developed a neural network as a simulation meta-model in economic analysis of risky projects. Many of the researches conducted on application of ANNs in assessment of high-tech projects' risk have more focused on approximating the value of the success or risk of the project while this paper's main assumption is that "project will be either successful or failed". So, in this paper, a model is proposed for recognizing the success or failure pattern of investing on high-tech projects.

**2.1. Technology Classifications**

Technologies can be studied in terms of various types (Aunger, 2010). As a matter of fact, there are some criteria based on which technologies can be classified into some types. Such a classification is represented in Table 2 (Aarabi and Mennati, 2014).

Table 1: ANNs' reported applications (Hakimpour et al. (2011))

| Business Area | Problem Type |
|---|---|
| Financial management and accounting | Financial health forecasting |
| | Assessment of compensation |
| | Classification of bankruptcy |
|  | Analytical inspection |
|  | Credit scoring and analysis |
|  | Signature verification analysis |
|  | Risk assessing |
|  | prediction |
|  | Classification of Stock trend |
|  | Bond evaluating and rating |
|  | Analysis of Interest rate |
|  | Selecting mutual found |
|  | Evaluation and rating of Credit |
| Sales and marketing | The response of costumers forecasting |
|  | Market development prediction |
|  | Sales forecast |
|  | Price elasticity modeling |
|  | Target marketing |
|  | Assessment of customer satisfaction |
|  | Customer loyalty and retention |
|  | Market segmentation |
|  | Analysis of customer behavior |
|  | Analysis of brand |
|  | Analysis of market basket |
|  | Storage layout study |
|  | Analysis of customer gender |
|  | Market orientation and performance |
|  | Study of marketing strategies, strategic planning and performance |
|  | Data mining in marketing |
|  | Prediction of marketing margin |
|  | New product adoption study |
|  | Forecasting of consumer choice |
|  | Approximation of market share |
| Production management | designing |
|  | Quality control applications |
|  | Planning and designing of Storage |
|  | Inventory controlling mechanism |
|  | Management of supply chain |
|  | Demand prediction |
|  | Monitoring and recognition |
|  | selection of process |
| Strategy and business study | Strategy and performance study |
|  | decision making assessment |
|  | Strategy evaluation |

Table 2: Technology types classification (Aarabi and Mennati, 2014).

| Criterion | Technology |
|---|---|
| Life Cycle | Emerging, Pacing, Key and basic Technologies |
| Labor or Capital | Labor and capital Intensive Technologies |
| Place | Intramural and extramural technologies |
| Complexity | Absorbable & non absorbable technologies |
| Output | High-tech, Medium Tech, Low Tech, labor-intensive technologies |
| Nature | Software & hardware technologies |
| Codification | Codified & Tacit technologies |
| background | Current and new technologies |
| Area use | Product and Process technologies |
| Appropriateness | Appropriate and inappropriate technologies |
| Importance | Critical /distinctive, basic and external technologies |

Development of High-tech projects needs both a lot of financial resources and too much supervision time. Moreover, investment of such projects entails a lot of risk and can't certainly lead to success. Therefore, some organizations have suffered enormous resource losses in process of investing on such projects because of the ignorance of risk assessment or using improper assessment methods (Jiang *et al*, 2010)

## 3. Development of a Risk Assessment Index System

To assess the risk of investing on high-tech projects, a Risk Assessment Index System (RAIS) should be developed at first. To do so, after interviewing some subject matter experts and studying related literature (Yongqing *et al*, 2009; Meredith *et al*, 2012; Song *et al*, 1999; Han *et al*, 2001 and Mao *et al*, 2002) Twenty-five variables related to the risk of high-tech project were captured and classified to six main risk contents as represented in Table 3. Then, the principal Component Analysis (PCA) was used to construct an index system. As a multivariate method, PCA has been widely used as an index construction method which reduces dimension by forming new variables (the principal components) as linear combinations of the variables in the multivariate set. The final result of using PCA to construct a RAIS from Table 3 is presented in Table 4.

Table 3: Risk contents and their risk variables

| Risk Contents | Risk variables |
|---|---|
| A: R & D Risks | A1:The financial resources availability |
| | A2:Capable human resources |
| | A3:Knowledge resources |
| B: Technical Risks | B1:Technical Maturity |
| | B2:Technology substitutability |
| | B3:Technology advantage |
| C: Production Risks | C1:The standardization degree of the production tools |
| | C2:The standardization degree of the production process |
| | C3:The supply capability of the raw material |
| D: Marketing Risks | D1:Market prospects |

|  | D2:Substitute products |
|  | D3:The Product life cycles |
|  | D4:Product competitiveness |
|  | D5:Possibility of new entrants |
| E: Management Risks | E1:The degree of managers' technical competencies |
|  | E2:The maturity of Project management methods |
|  | E3:The scientific weights of decisions |
|  | E4:The quality of managers' behavior |
| F:Environmental Risks | F1:The quality of conformation to cultural norms |
|  | F2:The degree of governmental support |

Table 4: RAIS of high-tech project investment

| Risk Contents | Risk variables |
|---|---|
| A: R & D Risks | A1:The financial resources availability |
|  | A2:Capable human resources |
|  | A3:Knowledge resources |
| B: Technical Risks | B1:Technical Maturity |
|  | B3:Technology advantage |
| C: Production Risks | C1:The standardization degree of the production tools |
|  | C2:The standardization degree of the production process |
|  | C3:The supply capability of the raw material |
| D: Marketing Risks | D1:Market prospects |
|  | D2:Substitute products |
|  | D4:Product competitiveness |
|  | D5:Possibility of new entrants |
| E: Management Risks | E1:The degree of managers' technical competencies |
|  | E3:The scientific weights of decisions |
|  | E4:The quality of managers' behavior |
| F:Environmental Risks | F1:The quality of conformation to cultural norms |
|  | F2:The degree of governmental support |

## 4. Pharmaceutical Industry

Pharmaceutical industry as an industry of high-tech products (i.e. drugs) is chosen as the case study of this paper. To conduct the research, it was very necessary to build a systematic and reliable questionnaire based on RAIS presented in Table 4. After constructing the questionnaire, it was sent to twelve firms which were active in pharmaceutical industry. These firms which were directly engaged in developing drug (as a high-tech product) had a lot of recorded date about their past experiences in developing drug products. The questionnaire was justified to all firms' managers and distributed to them from February 14, 2015 to February 16, 2015. The due time of questionnaire's reception was set for 10 days later (i.e. February 26, 2014).

Among all twelve firms that received the questionnaire just ten of them responded to it up to the end of due time. Data analysis showed that since received data were not completely synchronic, they had to be segmented into four time periods in order to cover all firms' recorded data. Therefore, the recorded data of these firms have been segmented into four periods as shown in Table 5. The received date showed that firms have had a very different performance in terms of successful (S) or failed (F) high-tech projects. This is represented in Table 6. It can be easily seen the 43% of projects conducted through 2000 to 2002 have been failed. 26% of projects conducted through 2003 to 2006 have been failed, 25% of projects conducted through 2006 to 2009 have been failed, 14% of projects conducted through 2010 to 2013 have been failed. Failure trend indicates the performance of firms in high-tech products development management has become better period by period. This may be largely due to the ascending knowhow that they have accumulated over time. Among other interesting points that can be taken from Table 6 is that the firm 10 has the best in all periods which it may be mainly because of its different resources, especially its intellectual ones.

Table 5: Firms' recorded data

|  | Number of implemented projects based on different periods | | | | Sum |
|---|---|---|---|---|---|
|  | From 2000 to 2002 | From 2003 to 2006 | From 2006 to 2009 | From 2010 to 2013 | |
| **Firm 1** | 3 | 5 | 7 | 6 | 21 |
| **Firm 2** | 2 | 6 | 6 | 10 | 24 |
| **Firm 3** | 4 | 5 | 7 | 7 | 23 |
| **Firm 4** | 3 | 4 | 4 | 6 | 17 |
| **Firm 5** | 6 | 9 | 9 | 8 | 32 |
| **Firm 6** | 3 | 5 | 6 | 6 | 20 |
| **Firm 7** | 3 | 3 | 4 | 4 | 14 |
| **Firm 8** | 4 | 5 | 7 | 8 | 24 |
| **Firm 9** | 3 | 5 | 8 | 7 | 23 |
| **Firm 10** | 6 | 5 | 5 | 6 | 22 |
| **Sum** | 37 | 52 | 63 | 68 | 220 |

Table 6: Firms' recorded date in terms of success or failure

| | Number of implemented projects based on different periods | | | | | | | |
|---|---|---|---|---|---|---|---|---|
| | From 2000 to 2002 | | From 2003 to 2006 | | From 2006 to 2009 | | From 2010 to 2013 | |
| | S | F | S | F | S | F | S | F |
| **Firm 1** | 2 | 1 | 3 | 2 | 5 | 2 | 5 | 1 |
| **Firm 2** | 1 | 1 | 4 | 2 | 5 | 1 | 8 | 2 |
| **Firm 3** | 1 | 3 | 4 | 1 | 5 | 2 | 5 | 2 |
| **Firm 4** | 1 | 2 | 3 | 1 | 2 | 2 | 5 | 1 |
| **Firm 5** | 4 | 2 | 6 | 3 | 6 | 3 | 7 | 1 |
| **Firm 6** | 2 | 1 | 4 | 1 | 4 | 2 | 5 | 1 |
| **Firm 7** | 0 | 3 | 2 | 1 | 3 | 1 | 4 | 0 |
| **Firm 8** | 3 | 1 | 4 | 1 | 5 | 2 | 7 | 1 |
| **Firm 9** | 3 | 0 | 5 | 0 | 7 | 1 | 7 | 0 |
| **Firm 10** | 4 | 2 | 3 | 2 | 4 | 1 | 5 | 1 |
| **Sum** | 21 | 16 | 38 | 14 | 47 | 16 | 58 | 10 |

## 5. Model Development

### 5.1. Artificial Neural Network

The ANN developed in this paper is represented in Figure 1:

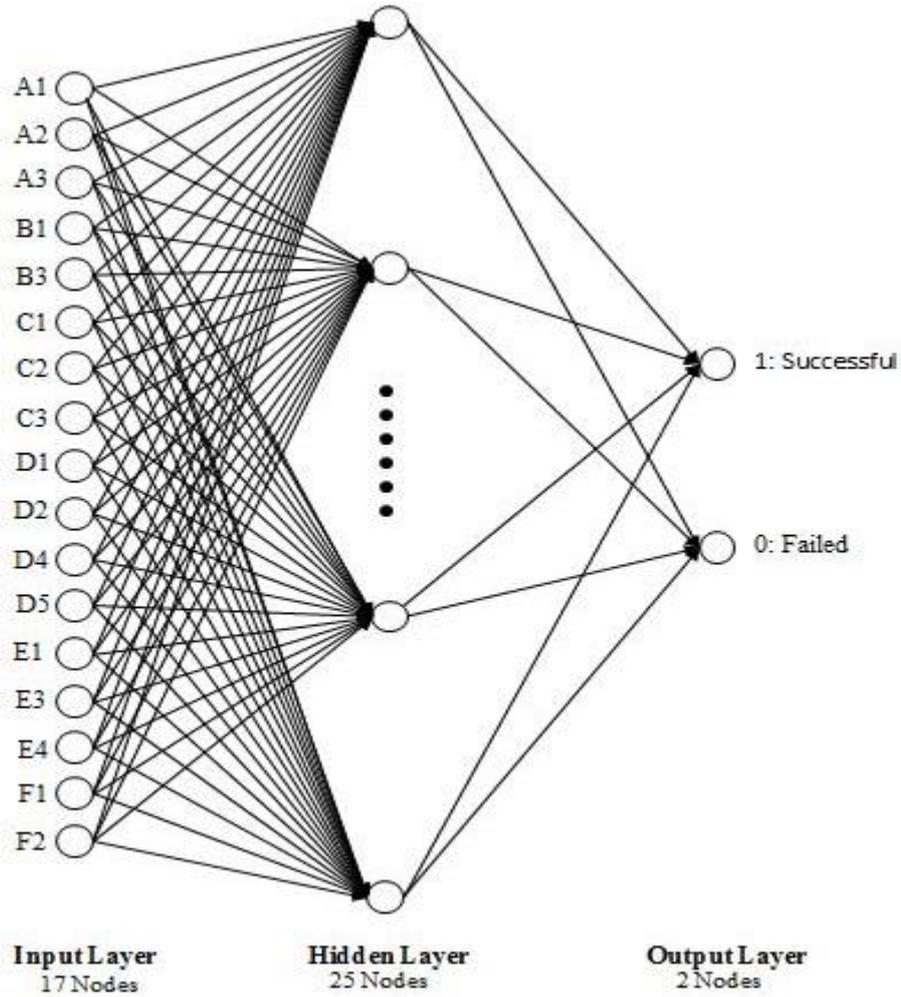

Figure 1: Proposed ANN

All input vectors of proposed ANN have 17 elements of RAIS. The number of these input vectors is equal to that of implemented projects (i.e. 220). The proposed ANN has 25 neurons (i.e. nodes) in its hidden layers each of which has a hyperbolic tangent sigmoid transfer function as follows:

$$f(n) = \frac{e^n - e^{-n}}{e^n + e^{-n}} \qquad (1)$$

Mathematically, it compresses all of its inputs to a range from -1 to +1, as it is shown in Figure 3 for an interval of [-10, 10].

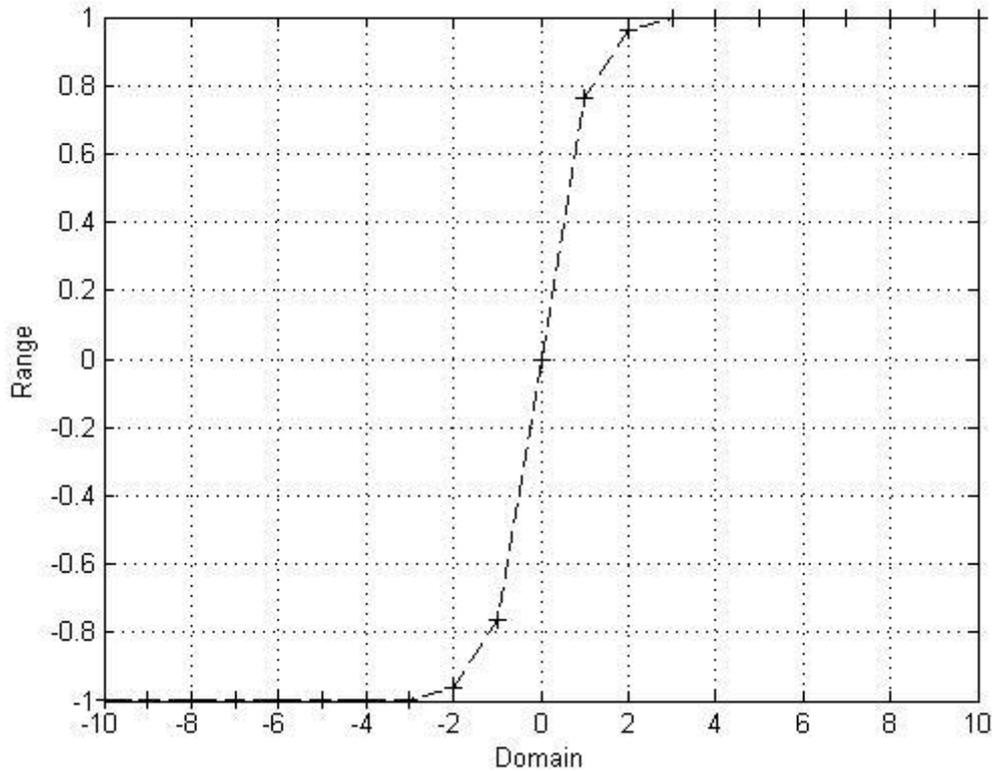

Figure 2: Function's diagram

In the output layer there are just 2 neurons equal to the number of classes (i.e. successful or failed). The performance function of proposed model is

$$MSE = \frac{\sum_{i=1}^{N} e_i^2}{N} \quad (2)$$

Actually, the model's most important purpose is to reduce this performance function as much as possible. To do so, back propagation algorithm has been found very efficient. This algorithm enables the network to update its parameters (i.e. weights and biases) in order to reduce performance function value. Parameter updating in turn is done through an iteratively training manner. Specifically, this mechanism enables the network to determine the gradient of performance function and by means of its training function updates its parameters for reducing performance function.

### 5.1.1. Model's Back Propagation Algorithm

As clearly explained by Hagan et al. (1996), in a multilayer network the output of one layer becomes an input for following one. This operation can be described by

$$a^{m+1} = f^{m+1}(W^{m+1}a^m + b^{m+1}) \quad m = 0,1,...,M-1 \quad (3)$$

Where M represents network's number of layers, first layer's neurons get external inputs

$$a^0 = p \quad (4)$$

by which the starting point is provided for equation (3). The outputs of the last layer's neurons are considered as network's outputs:

$$a = a^M \quad (5)$$

Mean Squared Error (MSE) as the performance function is used by back propagation. A set of examples of network's proper behavior are provided for it:

$$\{p_1,t_1\},\{p_2,p_1\},......,\{p_Q,t_Q\} \quad (6)$$

Where $p_q$ represents a network input and the corresponding target output is indicated by $t_q$. When each input enters the network, the network's output is compared with its target and in this case, algorithm has to adjust the parameters of network to minimize the value of MSE.

$$F(x) = E[e^2] = E[(t-a)^2] \quad (7)$$

Where x represents the vector of weights and biases of network and if the network has a number of outputs, this can be generalized to

$$F(x) = E[e^T e] = E[(t-a)^T(t-a)] \quad (8)$$

Then, it will approximate the MSE by

$$\hat{F}(x) = (t(k)-a(k))^T(t(k)-a(k)) = e^T(k)e(k) \quad (9)$$

Where the squared error at iteration k has replaced the expectation of the squared error. For calculating the steepest algorithm that can be used for calculating the approximate MSE is:

$$w_{i,j}^m(k+1) = w_{i,j}^m(k) - \alpha \frac{\partial \hat{F}}{\partial w_{i,j}^m} \qquad (10)$$

$$b_i^m(k+1) = b_i^m(k) - \alpha \frac{\partial \hat{F}}{\partial b_i^m} \qquad (11)$$

Where the $\alpha$ indicates the rate of learning

Partial derivatives of Eq. (10) and Eq. (11) can now be calculated by method of chain rule.

$$\frac{\partial \hat{F}}{\partial w_{i,j}^m} = \frac{\partial \hat{F}}{\partial n_i^m} \times \frac{\partial n_i^m}{\partial w_{i,j}^m} \qquad (12)$$

$$\frac{\partial \hat{F}}{\partial b_i^m} = \frac{\partial \hat{F}}{\partial n_i^m} \times \frac{\partial n_i^m}{\partial b_i^m} \qquad (13)$$

The calculation of each of above equations' second term can be easily done, because the net input to layer m is actually an explicit function of the parameters (weights and bias) in the layer:

$$n_i^m = \sum_{j=1}^{s^{m-1}} w_{i,j}^m a_j^{m-1} + b_i^m \qquad (14)$$

Therefore

$$\frac{\partial n_i^m}{\partial w_{i,j}^m} = a_j^{m-1}, \frac{\partial n_i^m}{\partial b_i^m} = 1 \qquad (15)$$

By defining the sensitivity of $\hat{F}$ for changes in the ith element of the net input in layer m

$$s_i^m = \frac{\partial \hat{F}}{\partial n_i^m} \tag{16}$$

Then Eq. (12) and Eq. (13) can be simplified to

$$\frac{\partial \hat{F}}{\partial w_{i,j}^m} = s_i^m a_j^{m-1} \tag{17}$$

$$\frac{\partial \hat{F}}{\partial b_i^m} = s_i^m \tag{18}$$

Now it is possible to represent the approximate steepest descent algorithm as

$$w_{i,j}^m(k+1) = w_{i,j}^m(k) - \alpha s_i^m a_j^{m-1} \tag{19}$$

$$b_i^m(k+1) = b_i^m(k) - \alpha s_i^m \tag{20}$$

Its matrix form can be represented as

$$w^m(k+1) = w^m(k) - \alpha s^m (a^{m-1})^T \tag{21}$$

$$b^m(k+1) = b^m(k) - \alpha s^m \tag{22}$$

Where

$$s^m = \frac{\partial \hat{F}}{\partial n^m} = \begin{bmatrix} \frac{\partial \hat{F}}{\partial n_1^m} \\ \frac{\partial \hat{F}}{\partial n_2^m} \\ \cdot \\ \cdot \\ \frac{\partial \hat{F}}{\partial n_m^m} \end{bmatrix} \tag{23}$$

Sensitivities have to be now calculated. For calculating $s^m$, the method of rule chain should be used again. This process is where the term of back propagation comes to surface, since a recurrence relationship in which the sensitivity at layer m is calculated from the sensitivity at layer m+1 is described.

If the recurrence relationship is going to be derived for the sensitivities, it is needed to use the following Jacobian Matrix:

$$\frac{\partial n^{m+1}}{\partial n^m} = \begin{bmatrix} \frac{\partial n_1^{m+1}}{\partial n_1^m} & \frac{\partial n_1^{m+1}}{\partial n_2^m} & \cdots & \frac{\partial n_1^{m+1}}{\partial n_{s^m}^m} \\ \frac{\partial n_2^{m+1}}{\partial n_2^m} & \frac{\partial n_2^{m+1}}{\partial n_2^m} & \cdots & \frac{\partial n_2^{m+1}}{\partial n_{s^m}^m} \\ \cdot & \cdot & & \cdot \\ \cdot & \cdot & & \cdot \\ \cdot & \cdot & & \cdot \\ \frac{\partial n_{s^{m+1}}^{m+1}}{\partial n_2^m} & \frac{\partial n_{s^{m+1}}^{m+1}}{\partial n_2^m} & \cdots & \frac{\partial n_{s^{m+1}}^{m+1}}{\partial n_{s^m}^m} \end{bmatrix} \quad (24)$$

If the i, j element of the above matrix is taken into account, it can be expressed as follows:

$$\frac{\partial n_i^{m+1}}{\partial n_j^m} = \frac{\partial \left( \sum_{t=1}^{s^m} w_{i,t}^{m+1} a_t^m + b_i^{m+1} \right)}{\partial n_j^m} = w_{i,j}^{m+1} \frac{\partial a_j^m}{\partial n_j^m} = w_{i,j}^{m+1} \frac{\partial f^m(n_j^m)}{\partial n_j^m} = w_{i,j}^{m+1} f^m(n_j^m) \quad (25)$$

Where

Therefore, the Jacobian matrix has to be written as

$$f^m(n_j^m) = \frac{\partial f^m(n_j^m)}{\partial n_j^m} \quad (26)$$

$$\frac{\partial n^{m+1}}{\partial n^m} = W^{m+1}\dot{F}^m(n^m) \tag{27}$$

When

$$\dot{F}^m(n^m) = \begin{bmatrix} \dot{f}^m(n_1^m) & 0 & 0 \\ 0 & \dot{f}^m(n_2^m) & 0 \\ 0 & 0 & \dot{f}^m(n_{s^m}^m) \end{bmatrix} \tag{28}$$

When chain rule is used in matrix form, the recurrence relation for sensitivity cab be written as following:

$$s^m = \frac{\partial \hat{F}}{\partial n^m} = \left(\frac{\partial n^{m+1}}{\partial n^m}\right)^T \frac{\partial \hat{F}}{\partial n^{m+1}} = \dot{F}^m(n^m)(W^{m+1})^T \frac{\partial \hat{F}}{\partial n^{m+1}} \tag{29}$$
$$= \dot{F}^m(n^m)(W^{m+1})^T s^{m+1}$$

In order to complete back propagation process, the starting point $s^m$ is required for the recurrence relation f Eq. (29) which is attained at the final layer:

$$s_i^M = \frac{\partial \hat{F}}{\partial n_i^M} = \frac{\partial (t-a)^T(t-a)}{\partial n_i^M} = \frac{\partial \sum_{j=1}^{S^M}(t_j - a_j)^2}{\partial n_i^M} = -2(t_i - a_i)\frac{\partial a_i}{\partial n_i^M} \tag{30}$$

And, because

$$\frac{\partial a_i}{\partial n_i^M} = \frac{\partial a_i^M}{\partial n_i^M} = \frac{\partial f^M(n_i^M)}{\partial n_i^M} = \dot{f}^M(n_i^M) \tag{31}$$

It is able to be written as

$$s_i^M = -2(t_i - a_i)\dot{f}^M(n_i^M) \tag{32}$$

That its matrix expression is

$$s^M = -2\dot{F}^M(n^M)(t-a) \tag{33}$$

Most often, BPs use a gradient descent algorithm for adjusting network's parameters. However, when the dimensions of ANNs get larger and more complicated, the Levenberg-Marquardt algorithm is strongly recommended especially because of its operation speed and accuracy. So, in this paper, a Levenberg-Marquardt back propagation has been used for network training.

## 5.2. Results Analysis

After writing and solving the proposed model by MATLAB Software, a set of various results was achieved. All of these results are presented as following:

### 5.2.1. Performance results

The network's total performance was calculated as 0.1782 which is really good. The other performances are shown in Figure 3.

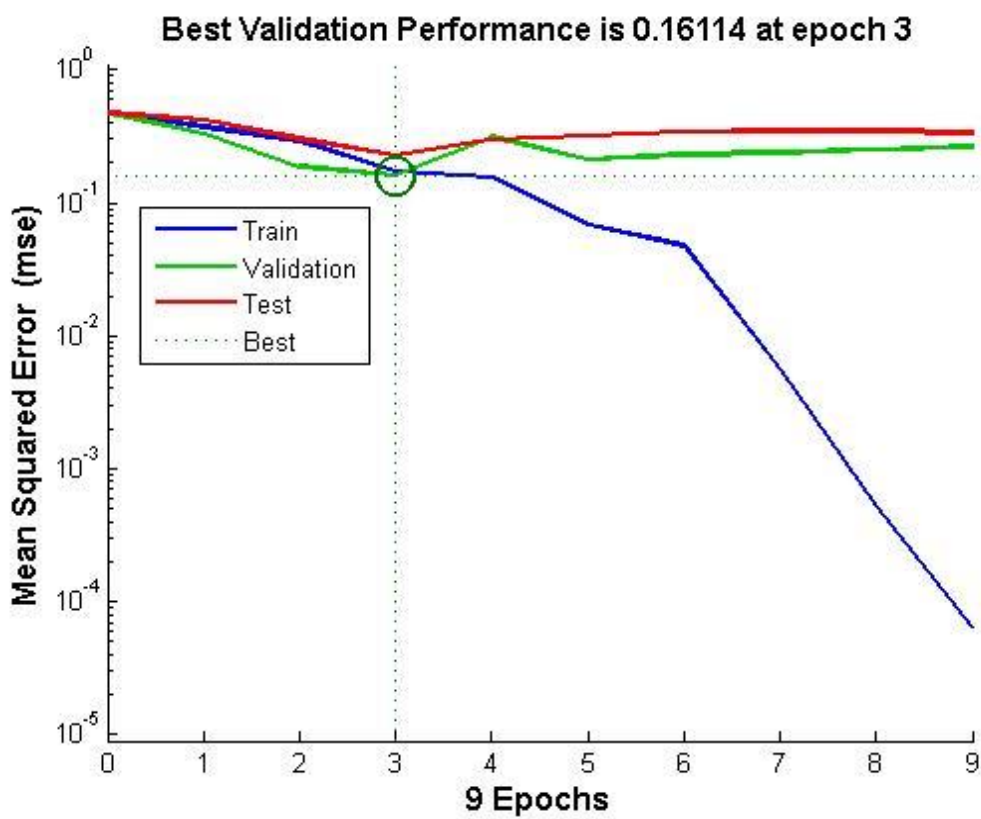

Figure 3: Network performances

The number of epochs is equal to the number of times that ANN has been allowed to be trained. As it is seen, all performances have had a descending order up to the third epoch.

Validation is the most important indicator for analyzing the network behavior. Actually, when this performance value goes up, it means that the network has started being over trained so its behavior will become unstable or chaotic over time. Therefore, the less validation performance value is, the more stable network's behavior is expected. However, while starting being over trained, the network training operation is stopped where the validation performance has had the least MSE value. As shown in Figure 3, the third epoch is where the network training operation has been stopped because after this point as shown in Figure 4, the network has reached maximum level of allowed failures (i.e. 6 failures).

However, the best performance value of this model is 0.1611 showing that the network behavior is really stable and its generalizability is very high.

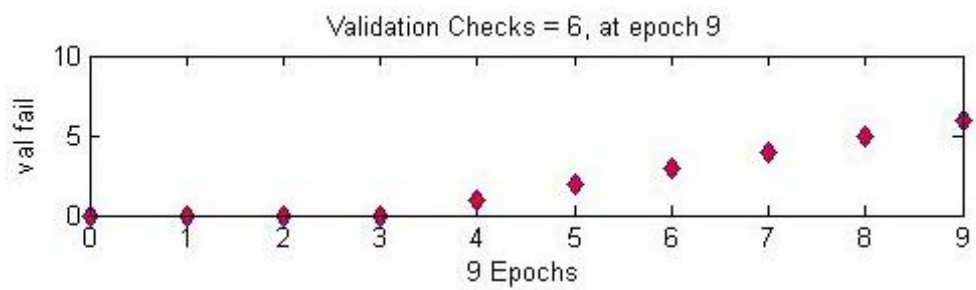

Figure 4: Validation failures

As an indicator for network training quality, training performance best value in the third epoch (i.e. 0.1720) shows that the network's training quality is really good in a way that its performance has become better in each of following epochs. The test performance which indicates network's learning quality is 0.2243 in the third epoch, this value means that ANN has had more errors in this performance index than other ones. However, this performance value is acceptable and proves ANN's good learning quality.

**5.2.2. Error Histogram**

Error Histogram of an ANN provides much precious information about its errors. Error Value (EV) and Error Frequency (EF) are two main data that can be extracted from error histogram. The variance of errors also shows that errors can be classified to big and small one in terms of Error Value. The negative sign of an error for each performance index happens when its outputs are larger than its targets. As shown in Figure 4, in training data set which entails 70% of all samples, most of errors are closed to zero (small errors) while the most of errors in test data set (which includes 15% of all samples) are far from zero (big errors).

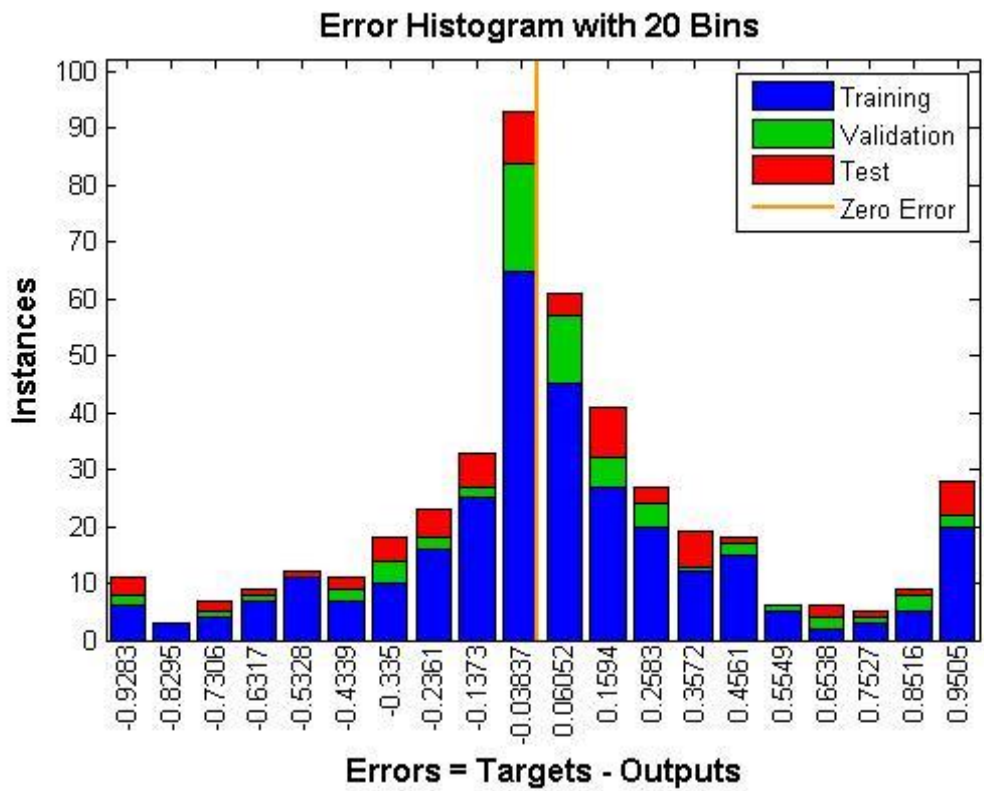

Figure 5: Error histogram

Errors of validation data set which entails 15% of all input samples are more inclined to zero meaning that the proposed ANN has a high degree of generalizability.

**5.2.3. Confusion Matrix**

Confusion matrices provide a lot of information about the precision and accuracy of network's results. Four types of confusion matrix have been presented in Figure 6. The training confusion matrix indicates that 119 of all samples allocated to training data set have been correctly classified and only 35 of them are misclassified. In other words, the proposed model can classify its training samples with accuracy of 77.3%.

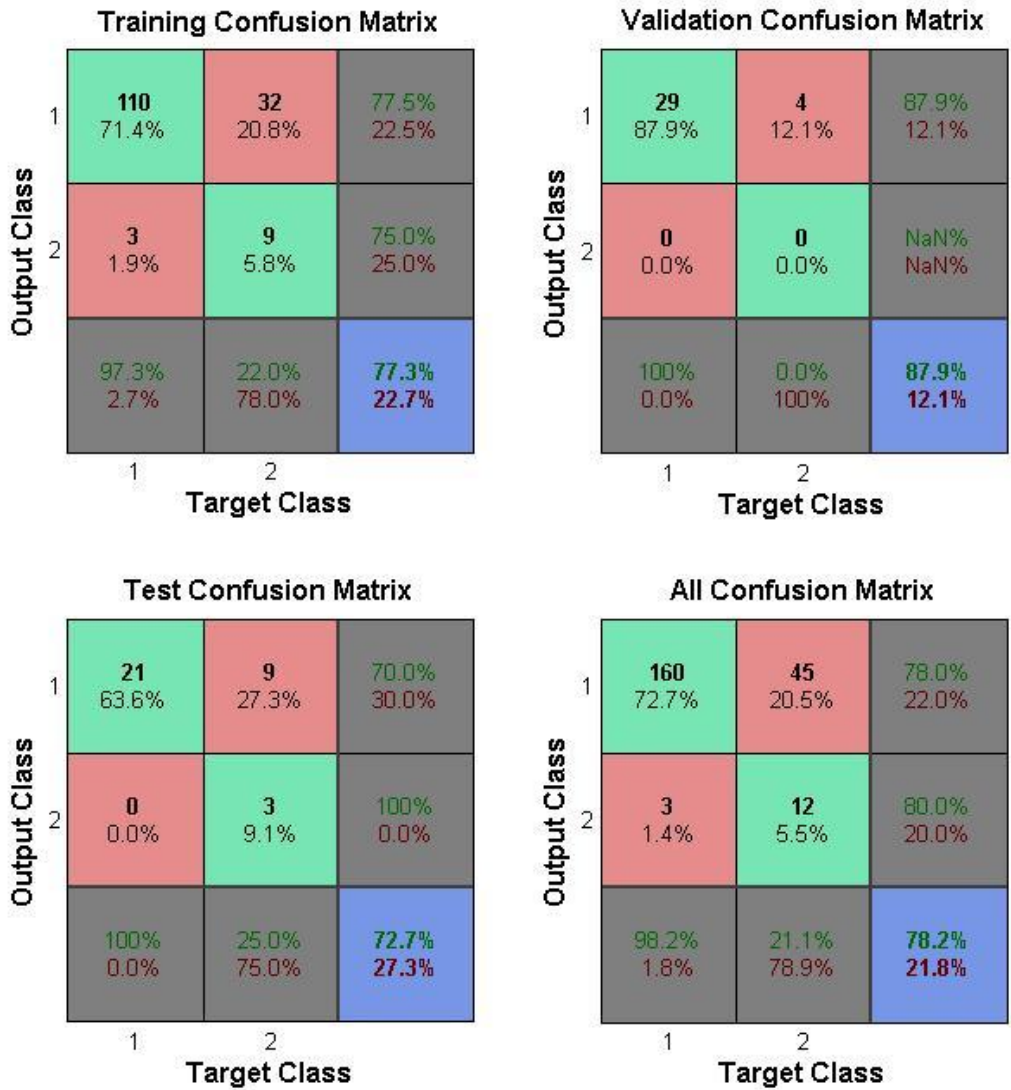

Figure 6: Four types of confusion matrix

The test confusion matrix indicates that 24 of all samples allocated to test data set have been correctly classified and only 9 of them are misclassified. In other words, the proposed model can classify its test samples with accuracy of 72.7%. The validation confusion matrix indicates that 29 of all samples allocated to test data set have been correctly classified and only 4 of them are misclassified. It other words, the proposed model can classify its validation samples with accuracy of 87.9%. It means that the network is highly generalizable and can be relied for decision making over time. The all confusion matrix represents the overall performance of proposed ANN in terms of classification accuracy. As it can be clearly seen, the model has been able to classify its samples with an accuracy of 78.2%.

## 5.3. Conclusion

Investing on high-tech products doesn't always yield the predicted results and organizations will suffer massive losses if their efforts in developing high-tech projects fail. To manage high-tech product development projects more confidently, managers need to have reliable information about their risk values in advance. The ANN proposed in this paper is aimed at helping managers to have such a precious information. Based on a Risk Assessment Index System (RAIS) that has been extracted from valid resources and constructed by Principal Component Analysis (PCA) method, an Artificial Neural Network (ANN) has been designed for enabling project managers to recognize the success or failure of each high-tech project before starting investing on it. The heighted level of model's accuracy and reliability makes it a very reliable mechanism for recognizing the success or failure of high-tech projects.

However, the proposed model can be improved in three aspects. The first aspect is about the methods by which researchers can enhance the performance of ANN's training function. Researchers such as Porto *et al*. (1995), Curry *et al*. (1997), Gupta *et al*. (1999), and Sexton *et al*. (2000) and Das *et al*. (2014) have studied on how decision makers can improve the training function of ANNs. The second aspect is that when there are many input variables (elements), it becomes painstakingly difficult to include all of them into the model. So, a mechanism should be developed for selecting more important input variables before they enter the model. Meta heuristics such as Genetic Algorithm (GA) and Ant Colony Optimization (ACO) can be used for doing so (see Das *et* *al*, 2014; Oreski and Oreski, 2014 and Monirul Kabir *et al*, 2012 and Sivagaminathan and Ramakrishnan, 2007). The third and last aspect is about the nature of model's variables which all can be dealt with in a fuzzy manner; therefore, development of a fuzzy ANN is strongly needed (see Chien *et al*, 2010 and Ku, 2001). Anyway, pursuing each of these three aspects is of paramount value and can be a subject for future researches.